\documentclass{article}

\usepackage{arxiv}

\usepackage[utf8]{inputenc} 
\usepackage[T1]{fontenc}    
\usepackage{hyperref}       
\usepackage{url}            
\usepackage{booktabs}       
\usepackage{amsfonts}       
\usepackage{nicefrac}       
\usepackage{microtype}      
\usepackage{lipsum}
\usepackage{graphicx}
\graphicspath{ {./images/} }
\usepackage{graphicx}
\usepackage{amssymb}
\usepackage{amsmath}
\usepackage{float}
\usepackage{placeins}
\usepackage[booktabs]{}
\title{MentalMARBERT: Domain-Adaptive Pre-training and Two-Stage Fine-Tuning
for Arabic Mental Health Disorders Detection}
\author{
\begin{tabular}{cc}
\begin{tabular}{c}
\textbf{Fatimah Almalki}\\
\textnormal{Department of Computer Science}\\
\textnormal{Faculty of Computing and Information Technology}\\
\textnormal{King Abdulaziz University}\\
\textnormal{Jeddah, Saudi Arabia}\\
\texttt{falmalki0402@stu.kau.edu.sa}
\end{tabular}
&
\begin{tabular}{c}
\textbf{Areej Alhothali}\\
\textnormal{Department of Computer Science}\\
\textnormal{Faculty of Computing and Information Technology}\\
\textnormal{King Abdulaziz University}\\
\textnormal{Jeddah, Saudi Arabia}\\
\texttt{aalhothali@kau.edu.sa}
\end{tabular}
\\[1.2cm]
\begin{tabular}{c}
\textbf{Lulwah Alharigy}\\
\textnormal{Department of Computer Science}\\
\textnormal{Faculty of Computing and Information Technology}\\
\textnormal{King Abdulaziz University}\\
\textnormal{Jeddah, Saudi Arabia}\\
\texttt{lalharigy@kau.edu.sa}
\end{tabular}
&
\begin{tabular}{c}
\textbf{Abdulrahman Aladeem}\\
\textnormal{Department of Psychology}\\
\textnormal{College of Arts and Humanities}\\
\textnormal{King Abdulaziz University}\\
\textnormal{Jeddah, Saudi Arabia}\\
\texttt{Psy.a.aladeem@outlook.com}
\end{tabular}
\end{tabular}
}

\begin{document}
\maketitle
\begin{abstract}
Detecting mental health disorders from Arabic social media text remains challenging due to dialectal variation, informal language, limited high-quality annotated resources, and severe class imbalance. While English mental health natural language processing (NLP) has progressed substantially, Arabic multi-class disorder classification remains insufficiently studied. This study proposes a two-phase framework for Arabic mental health text classification. In phase 1, three Arabic pre-trained language models, AraBERT, CAMeLBERT, and MARBERT, undergo Domain-Adaptive and Task-Adaptive Pretraining (DAPT and TAPT) using a large-scale corpus of unlabeled Arabic mental health tweets. The adapted models are evaluated under a unified protocol to identify the most effective backbone model. In phase 2, the selected model is assessed across four configurations combining single-stage and hierarchical two-stage classification architectures with Full Fine-Tuning and Low-Rank Adaptation (LoRA). To support this study, we constructed a novel annotated Arabic mental health dataset comprising 50,670 tweets across six categories, with strong inter annotator agreement (Krippendorff’s Alpha = 0.733, average pairwise agreement = 0.797). Experimental results show that the domain-adapted MARBERT (MentalMARBERT) achieves statistically significant improvements over baseline models in both accuracy and macro-F1. The hierarchical two-stage architecture combined with full fine-tuning achieves the best overall performance, reaching a macro-F1 of 0.861 and an accuracy of 0.877. These findings demonstrate the effectiveness of domain-specific adaptive pretraining and hierarchical classification for Arabic mental health disorder detection.
\end{abstract}


\section{Introduction}
Mental health is a fundamental pillar of both personal and collective well-being, playing a significant role in shaping individuals’ quality of life, their productive capacity, and the stability of society as a whole. The World Health Organization (WHO) reports that over one billion individuals across the globe are currently affected by mental health conditions, positioning such disorders among the foremost contributors to disability worldwide~\cite{WHO2025}. Disorders such as depression, anxiety, and post-traumatic stress disorder (PTSD) contribute substantially to disability-adjusted life years (DALYs) and years lived with disability (YLDs), particularly in regions affected by conflict and sociopolitical instability, including parts of the Middle East and North Africa (MENA)~\cite{Effatpanah2023}~\cite{AlGarni2025}. Despite the growing burden, delayed diagnosis and limited access to mental health services remain persistent challenges, especially in Arab societies where stigma and limited public awareness often hinder timely intervention~\cite{Antoun2020}. Traditional diagnostic approaches rely primarily on clinical assessments and self-reported questionnaires, which can be time-consuming and inaccessible to large segments of the population. In recent years, advances in natural language processing (NLP) and artificial intelligence (AI) have introduced new opportunities for large-scale mental health monitoring through the analysis of linguistic patterns in social media text\cite{Tounsi2025}. Transformer-based large language models (LLMs) have demonstrated strong performance across multiple text classification tasks, including mental health detection\cite{Zhang2022}\cite{Pandey2024MentalHA}. However, individuals frequently express psychological distress using culturally specific, indirect, and dialectal language, posing challenges for general-purpose models trained on broad-domain corpora. To address domain-specific limitations, prior research has explored specialized mental health models in English, demonstrating that domain-adaptive pre-training improves performance by enhancing contextual representations of disorder-related language \cite{Zhang2022}. Nevertheless, Arabic mental health NLP remains relatively underexplored~\cite{ji2022mentalbert}. Existing studies have primarily focused on binary detection of depression or suicidality, often using general-domain Arabic models, such as AraBERT or MARBERT without systematic domain adaptation\cite{Abdul-mageed2021} Furthermore, Arabic presents additional challenges, including substantial dialectal variation, informal orthography, and culturally specific expressions that are often underrepresented in pre-training corpora \cite{Antoun2020}. These factors highlight the need for domain-adapted Arabic models and robust evaluation frameworks tailored to multi-class mental disorder classification~\cite{Thakkar}.
To bridge this gap, this study proposes a comprehensive two-phase framework for the classification of Arabic mental health texts. In the first phase, three Arabic pre-trained language models AraBERT, CAMeLBERT, and MARBERT undergo domain-adaptive and task-adaptive pre-training (DAPT and TAPT) using a large-scale corpus of unlabeled Arabic mental health tweets to identify the most suitable backbone model. In the second phase, the best-performing model is evaluated across multiple classification architectures, including single-stage and hierarchical two-stage designs, as well as different fine-tuning strategies, namely full fine-tuning and Low-Rank Adaptation (LoRA).
To support this investigation, we constructed a novel expert-annotated Arabic mental health dataset comprising 50,670 tweets across six categories: None, Depression, Anxiety, Bipolar Disorder, {Post-Traumatic Stress Disorder (PTSD), and Obsessive-Compulsive Disorder (OCD). To the best of our knowledge, this is the first comprehensive study that jointly evaluates domain-adaptive pre-training, hierarchical classification architectures, and parameter-efficient fine-tuning for multi-class Arabic mental health disorder detection using a large scale expert-annotated dataset.

The main contributions of this study are summarized as follows. First, we propose a two-phase framework for Arabic mental health disorder classification that integrates domain-adaptive pre-training, hierarchical classification architectures, and parameter-efficient fine-tuning strategies. Second, we construct a large-scale expert-annotated Arabic mental health dataset consisting of 50,670 tweets across six categories (None, Depression, Anxiety, Bipolar Disorder, PTSD, and OCD). Third, we perform systematic domain-adaptive and task-adaptive pre-training (DAPT and TAPT) on three Arabic pre-trained language models (AraBERT, CAMeLBERT, and MARBERT) to identify the most effective backbone model. Finally, we evaluate different classification architectures and adaptation strategies, including full fine-tuning and Low-Rank Adaptation (LoRA).

\section {Related Work} 
A wide range of classification algorithms has been proposed to distinguish textual data associated with individuals experiencing mental health conditions from that of unaffected individuals~\cite{Lorenzoni2024}. In order to present a well-organized summary of existing literature, this section is divided into three parts. The first part covers conventional machine learning techniques, while the second explores deep learning approaches, with particular emphasis on transformer-based models such as BERT and RoBERTa, which have emerged as key drivers of progress in mental health-related NLP. The third part addresses large language models that have been specifically customized and adapted to the domain of mental health NLP
\subsection {Machine Learning Models}

Jaman et al.~\cite{Jaman2021} aimed to classify several mental health disorders, including schizophrenia, PTSD, bipolar disorder, and depression, by analyzing text extracted from Reddit posts. Their study employed multiple machine learning classifiers namely SVM, Logistic Regression, GRU, and BERT to detect indicators of these conditions. Each model was evaluated against a control group for every disorder. Results showed that SVM achieved the highest accuracy of 87\% for PTSD detection, whereas Logistic Regression attained 82\% accuracy in classifying bipolar disorder.
Alzoubi et al.~\cite{Alzoubi2024Depression} gathered a dataset of Arabic tweets to investigate the automated detection of depressive symptoms through machine learning. Among the tested algorithms, Mutational Naïve Bayes combined with TF-IDF outperformed the rest, achieving an accuracy of 86\% in categorizing tweets as either depressive or non-depressive.
Lorenzoni et al.~\cite{Lorenzoni2024} conducted a comparative evaluation of several ML and NLP methods using the DAIC-WOZ dataset a collection of clinical interviews designed for diagnosing depression, anxiety, and PTSD with a specific focus on depression detection. Three classifiers were implemented and benchmarked: Random Forest, XGBoost, and SVM. The findings indicated that Random Forest and XGBoost delivered the strongest results, both achieving approximately 84\ accuracy, while SVM lagged behind at around 72\%.
\subsection {Deep Learning and Transformer-Based Models} 

Qayyum et al.~\cite{Qayyum2023} proposed a framework for diagnosing mental health conditions using Reddit data, covering six disorders: depression, anxiety, bipolar disorder, schizophrenia, autism, and general mental health. The study evaluated a range of models including CNN, LSTM, GRU, bidirectional GRU, BERT, and RoBERTa. A hierarchical RoBERTa-based framework that organizes disorders into a taxonomy was introduced, achieving an accuracy of 84\%.
Arif et al.~\cite{Arif2024} developed a multi-class classification model for detecting common mental illnesses from Reddit posts. Deep learning architectures such as CNN, LSTM, and Bi-LSTM were compared alongside transfer learning models including BERT, XLNet, and RoBERTa. Among all evaluated models, RoBERTa attained the highest overall accuracy of 83\%.
Xu et al.~\cite{Xu2024} examined the capability of LLMs in predicting mental health conditions from online text, evaluating models such as FLAN-T5, GPT-3.5, GPT-4, and Alpaca across stress, depression, and suicide risk prediction tasks. Fine-tuned models consistently outperformed their zero-shot counterparts.
Hassan et al.~\cite{Hassan2024} introduced an automated multi-label annotation methodology for mental health disorders using LLMs, alongside the SPAADE-DR dataset designed to capture co-occurring conditions. Among the evaluated models, Llama-3 70B achieved the highest balanced accuracy of 78\% in multi-label classification. Elmajali et al.~\cite{Elmajali2024} investigated early depression detection in Arabic tweets using AraBERT and MARBERT for multi-class symptom classification. AraBERT delivered the best performance, achieving an accuracy of 99.3\% and an F1-score of 98.9\%. Ilias et al.~\cite{Ilias2024} proposed transformer-based models enhanced with extra-linguistic features, including sentiment lexicons and linguistic markers, for stress and depression detection in social media. The approach achieved an F1-score of 83.10\% for depression classification. Pourkeyvan et al.~\cite{Pourkeyvan} assessed the performance of several Hugging Face pre-trained BERT variants including DistilBERT, BERT-base-uncased, MentalBERT, and DistilRoBERTa for mental health detection from Twitter data comprising over 11.8 million tweets and 553 user bios. DistilBERT demonstrated the strongest performance, reaching an accuracy of 97\%.

\subsection {Specialized Large Language Models for Mental Health}
Large Language Models (LLMs) have emerged as a pivotal tool in the field of mental health, specifically for the early detection and treatment of mental disorders using social media content~\cite{Hua2024}. Building on advances in Natural Language Processing (NLP), transformer-based encoder architectures such as BERT and its variants have been adapted to mental health-related tasks, demonstrating considerable promise in processing and understanding domain-specific textual data~\cite{ji2022mentalbert}. Nevertheless, the effectiveness of general-purpose models in this domain remains constrained by their reliance on broad pretraining corpora that lack mental health-specific knowledge. To overcome this limitation, researchers have developed specialized models tailored to the mental health domain, most notably MentalBERT, MentalRoBERTa, and Chinese-MentalBERT.
Ji et al.~\cite{ji2022mentalbert} introduced MentalBERT and MentalRoBERTa, two domain-specifically pretrained masked language models for the mental healthcare research community. The models were developed by applying Domain-Adaptive Pretraining (DAPT) to BERT-base and RoBERTa, using a large corpus of mental health-related posts collected from Reddit subreddits, including r/depression, r/SuicideWatch, and r/Anxiety. The models were subsequently fine-tuned and evaluated on eight downstream classification tasks covering depression detection (eRisk, CLPsych 2015, Depression Reddit), stress (Dreaddit), suicidal ideation (UMD Suicidality, T-SID), multi-label disorder detection (SWMH), and stressor classification (SAD). Results demonstrated that domain-adapted models consistently outperformed general-purpose models (BERT and RoBERTa) as well as biomedical (BioBERT) and clinical (ClinicalBERT) variants across most tasks. Specifically, MentalBERT achieved the highest F1 score on the UMD Suicidality dataset, while MentalRoBERTa led performance on the T-SID, Dreaddit, SAD, and SWMH benchmarks, demonstrating that continued pretraining on target-domain corpora, followed by task-specific fine-tuning, is an effective strategy for mental health disorder detection. Building on this direction, Zhai et al.~\cite{Zhai2023} developed Chinese-MentalBERT to address the absence of domain-adapted mental health models for non-English languages. Using DAPT, the authors continued pretraining the Chinese-BERT-wwm-ext model on posts collected from Chinese social media platforms, enriched with publicly available mental health datasets. A key methodological contribution was the integration of a psychological lexicon-guided masking strategy during pretraining, which directed the model’s attention toward domain-relevant psychiatric vocabulary rather than relying on random token masking. The model was evaluated across six tasks, including emotion classification, cognitive distortion detection, and suicide risk assessment, outperforming eight baseline models, underscoring the value of culturally and linguistically adapted pretraining for mental health NLP. Taking a different approach, Tounsi et al.~\cite{Tounsi2025} proposed Ara-MentalBERT, a fusion model that combines AraBERT and MentalBERT through a Deep Joint Autoencoder (DJAE) to detect mental health disorders from Arabic social media. The model was evaluated on two datasets: an English Reddit dataset covering six mental health disorders and an Arabic Twitter dataset restricted to depression-related symptoms with approximately 1,221 samples. On the Reddit dataset, Ara-MentalBERT achieved 81.12\% accuracy, outperforming all baselines. However, on the Arabic Twitter dataset, AraBERT outperformed the proposed model (F1: 93.56\% vs. 85.78\%), suggesting that dedicated Arabic language models may hold an advantage on Arabic-only data. It should be noted that the Arabic dataset was restricted solely to depression-related symptoms, which may limit the generalizability of these findings.

\section {Dataset} 
In this study, two distinct datasets were constructed to support different training stages of the proposed framework for Arabic mental health disorder detection. Both datasets were collected from Twitter (X.com) using the Apify Twitter Search Scraper actor~\cite{Apify2025}, which provides automated capabilities for large-scale social media data extraction. Twitter was selected as the primary data source due to its widespread use among Arabic-speaking populations for expressing personal experiences and emotional states, making it a rich and representative source for mental health related discourse in Arabic. The overall dataset construction process is illustrated in Fig.~\ref{fig:dataset_pipeline}
\begin{figure}[t]
\centering
\includegraphics[width=0.5\columnwidth]{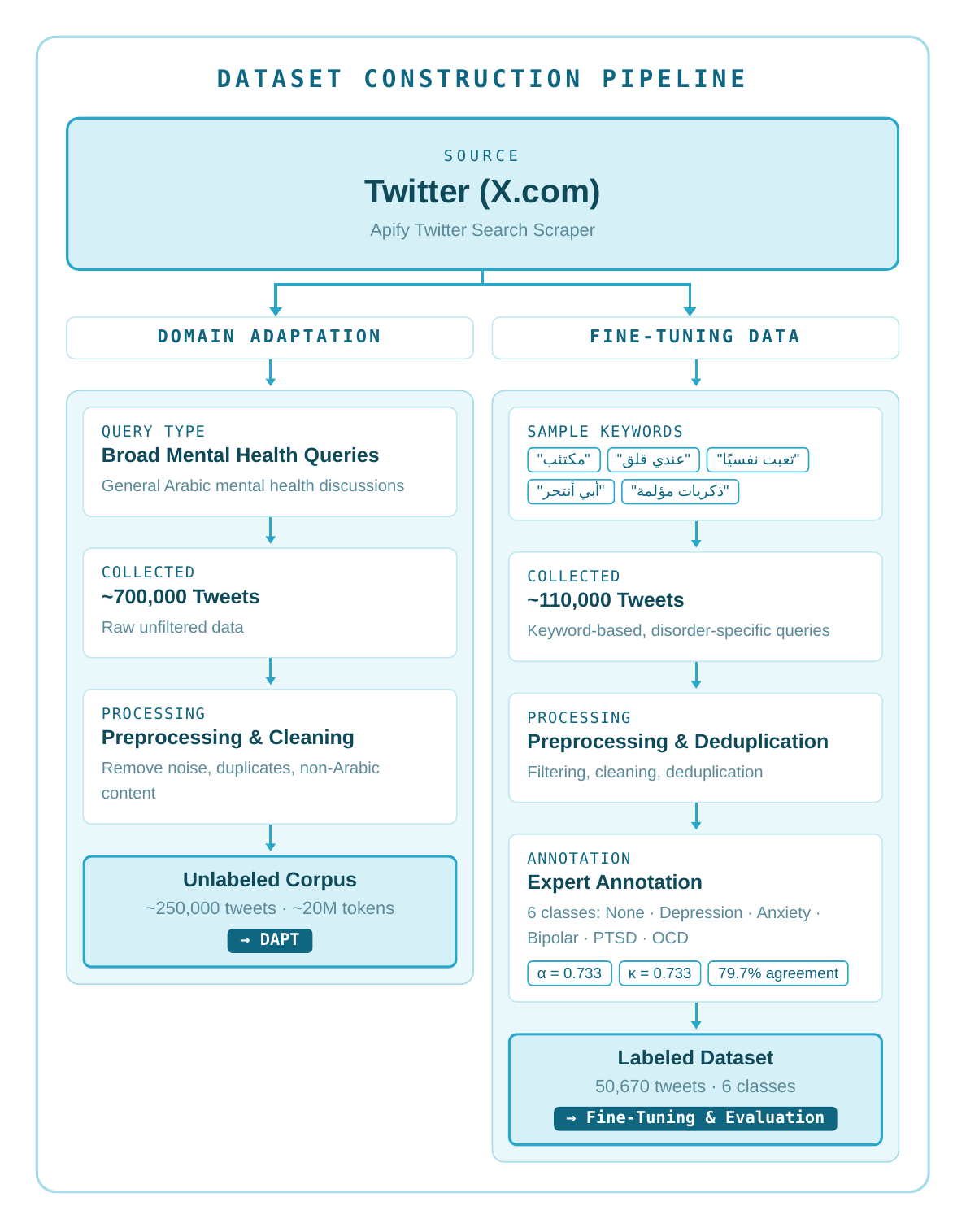}
\caption{Dataset pipeline}
\label{fig:dataset_pipeline}
\end{figure}

\subsection {Domain Adaptation Dataset}
The domain adaptation dataset was constructed to support continued pre-training of the base language models on Arabic mental health–related social media content. Broad mental health–related search queries were employed to capture diverse discussions relevant to psychological well-being in Arabic social media, resulting in an initial collection of approximately 700,000 tweets. The collected data underwent a series of preprocessing steps, including the removal of URLs, mentions, emojis, and non-Arabic content, as well as deduplication to eliminate redundant entries and normalization of Arabic text~\cite{Antoun2020}. Following preprocessing, the dataset was reduced to approximately 250,000 tweets, yielding a corpus of approximately 20 million tokens. This dataset consists entirely of unlabeled content and was used exclusively for domain-adaptive pre-training (DAPT) prior to supervised fine-tuning.

\subsection {Fine Tuning Dataset} 
The fine-tuning dataset was developed to support supervised multi-class classification of mental health related Arabic tweets. The six target classes: depression, anxiety, PTSD, bipolar disorder, OCD, and None were selected based on their prevalence in Arabic social media discourse and their clinical relevance as commonly reported mental health conditions in the Arabic-speaking population. Tweet retrieval relied on a predefined set of Arabic keywords derived from prior studies in the literature, corresponding to commonly used self-reported expressions in mental health–related social media discourse, including terms conveying meanings such as feeling depressed, experiencing anxiety, being mentally exhausted, expressing suicidal ideation, and recalling painful memories, among others. These keywords were used exclusively for data collection and were not employed as labeling criteria. An initial collection of approximately 110,000 tweets was gathered, which was subsequently subjected to the same preprocessing pipeline applied to the domain adaptation dataset, including removal of URLs, mentions, emojis, and non-Arabic content, deduplication, and text normalization. Following preprocessing, the final dataset consists of 50,670 tweets, as summarized in Table ~\ref{tab:disorder_distribution}.

\begin{table}[htbp]
\centering
\caption{Disorder Distribution in the Dataset} \label{tab:disorder_distribution}
\begin{tabular}{@{}lc@{}} 
\toprule
\textbf{Disorder} & \textbf{Number of Samples} \\
\midrule
None        & 16{,}416 \\
Depression  & 13{,}469 \\
Anxiety     & 10{,}073 \\
Bipolar     & 3{,}843  \\
PTSD        & 3{,}562  \\
OCD         & 3{,}307  \\
\midrule
\textbf{Total} & \textbf{50{,}670} \\
\bottomrule
\end{tabular}
\end{table}

As shown in Table~\ref{tab:disorder_distribution} and Fig~\ref{fig:disorder_distribution}, the dataset exhibits a notable class imbalance, with the None class representing the largest proportion of samples, while disorder-specific classes such as OCD, PTSD, and bipolar contain fewer samples. This imbalance motivated the adoption of Stratified K-Fold Cross-Validation and macro-F1 as the primary evaluation metric to ensure unbiased performance estimates across all classes.
\begin{figure}[htbp]
\centering
\includegraphics[width=0.7\columnwidth]{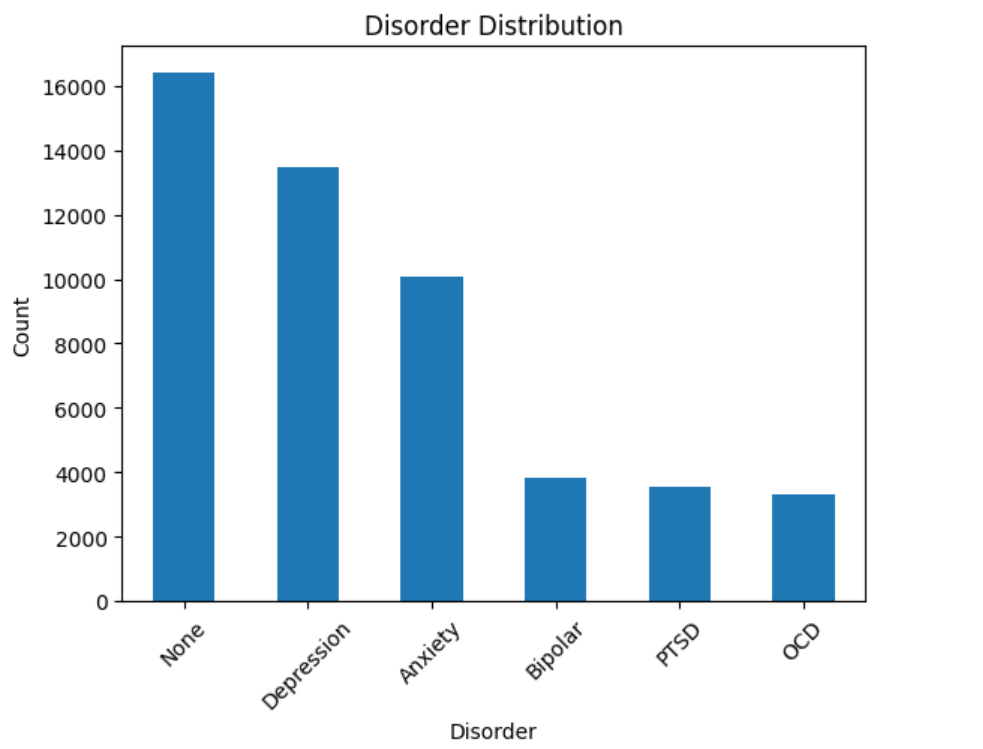}
\caption{Distribution of mental health disorder categories in the dataset}
\label{fig:disorder_distribution}
\end{figure}

\subsection {Data Annotation} 

The fine-tuning dataset was annotated for multi-class classification by four domain experts in psychology and mental health. Each tweet was independently assigned to one category from a predefined label set comprising \textit{Depression}, \textit{Anxiety}, \textit{Post-Traumatic Stress Disorder (PTSD)}, \textit{Bipolar Disorder}, \textit{Obsessive-Compulsive Disorder (OCD)}, and \textit{None}. Examples of annotated texts for each class are presented in Fig. ~\ref{fig:samples}.

\begin{figure}[H]
    \centering
    \includegraphics[width=\linewidth]{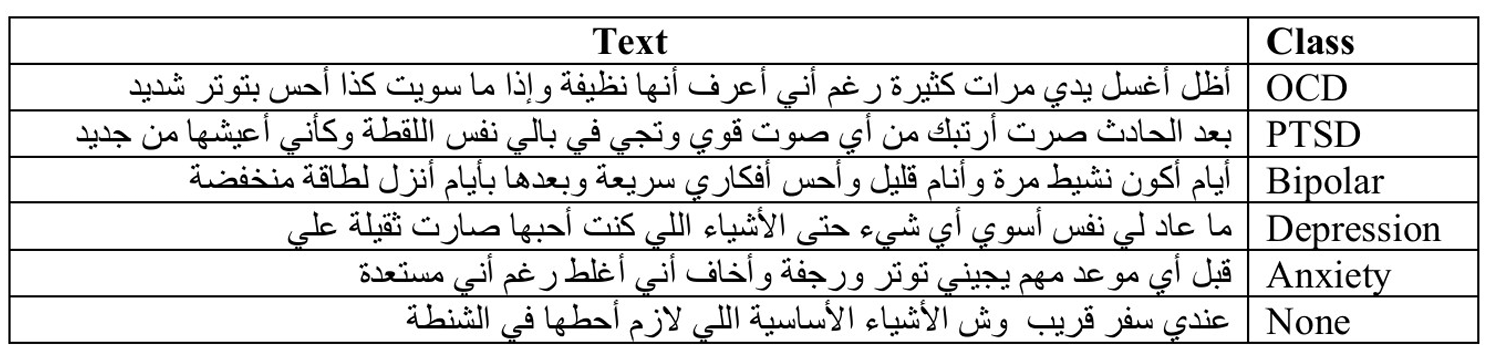}
    \caption{Samples of the labeled dataset.}
    \label{fig:samples}
\end{figure}
To ensure the reliability and consistency of the annotation process, inter-annotator agreement (IAA) was evaluated using three statistical measures, each capturing a different aspect of agreement to provide a more comprehensive and robust evaluation.

\subsubsection {Krippendorff's Alpha}

Krippendorff's Alpha ($\alpha$) is a generalized agreement coefficient applicable to any number of annotators and measurement scales~\cite{krippendorff2018}. It was selected in this study because it accommodates multiple annotators and is suitable for nominal classification tasks. It is defined as shown in Equation (1):

\begin{equation}
\alpha = 1 - \frac{D_o}{D_e}
\end{equation}

where $D_o$ represents the observed disagreement and $D_e$ denotes the expected disagreement by chance. A value of $\alpha = 1$ indicates perfect agreement, while $\alpha = 0$ corresponds to agreement equivalent to chance.

\subsubsection {Fleiss' Kappa}

Fleiss' Kappa ($\kappa$) measures agreement among multiple annotators for categorical labels~\cite{fleiss2003}. It was chosen as it is specifically designed for measuring agreement among three or more annotators on categorical labels. It is computed as shown in Equation (2):

\begin{equation}
\kappa = \frac{\bar{P} - \bar{P}_e}{1 - \bar{P}_e}
\end{equation}

where $\bar{P}$ is the mean observed agreement across all annotators and categories, and $\bar{P}_e$ is the expected agreement by chance.

\subsubsection {Average Pairwise Cohen's Kappa}
Cohen's kappa is a widely used measure for assessing pairwise agreement between two annotators~\cite{cohen1960}\cite{mchugh2012}. It was included to examine consistency between each pair of annotators individually, providing a more detailed view of agreement. It is calculated between each pair and then averaged, as shown in Equation (3):

\begin{equation}
\kappa = \frac{p_o - p_e}{1 - p_e}
\end{equation}

where $p_o$ denotes the observed agreement between two annotators and $p_e$ represents the expected agreement by chance.

According to commonly accepted interpretation guidelines~\cite{landis1977}, kappa values between 0.61 and 0.80 indicate substantial agreement. Therefore, the obtained values of Krippendorff's Alpha = 0.733, Fleiss' Kappa = 0.733, and an average pairwise percent agreement of approximately 0.797 as summarized in Table~\ref{tab:inter_annotator_agreement}. All fall within the substantial agreement range, confirming the quality and reliability of the labeled dataset.

\begin{table}[!ht]
\caption{Inter-Annotator Agreement Results}
\label{tab:inter_annotator_agreement}
\centering
\begin{tabular}{l c}
\hline
\textbf{Metric} & \textbf{Value} \\
\hline
Krippendorff’s Alpha (nominal) & 0.733 \\
Fleiss’ Kappa & 0.733 \\
Average Pairwise Agreement & 79.7\% \\
\hline
\end{tabular}
\end{table}
\section {Methodology} 
This section presents the proposed methodology for Arabic mental health text classification. The overall framework is organized into two sequential phases, as illustrated in Figs.~\ref{fig:phase1} and ~\ref{fig:phase2}. The first phase focuses on baseline model selection, where multiple Arabic pre-trained language models are evaluated under a unified and controlled experimental setup to determine the most robust and domain-adaptable model for the task. The second phase introduces the proposed classification framework, in which the best-performing model identified in Phase 1 is further evaluated across four experimental configurations combining two classification strategies, single-stage and two-stage classification, with two fine-tuning approaches: full fine-tuning (Full FT) and Low-Rank Adaptation (LoRA). This structured evaluation enables a comprehensive analysis of both architectural design choices and parameter-efficient adaptation strategies for Arabic mental health classification.

\begin{figure}[!htbp]
   \centering
    \includegraphics[width=0.5\linewidth]{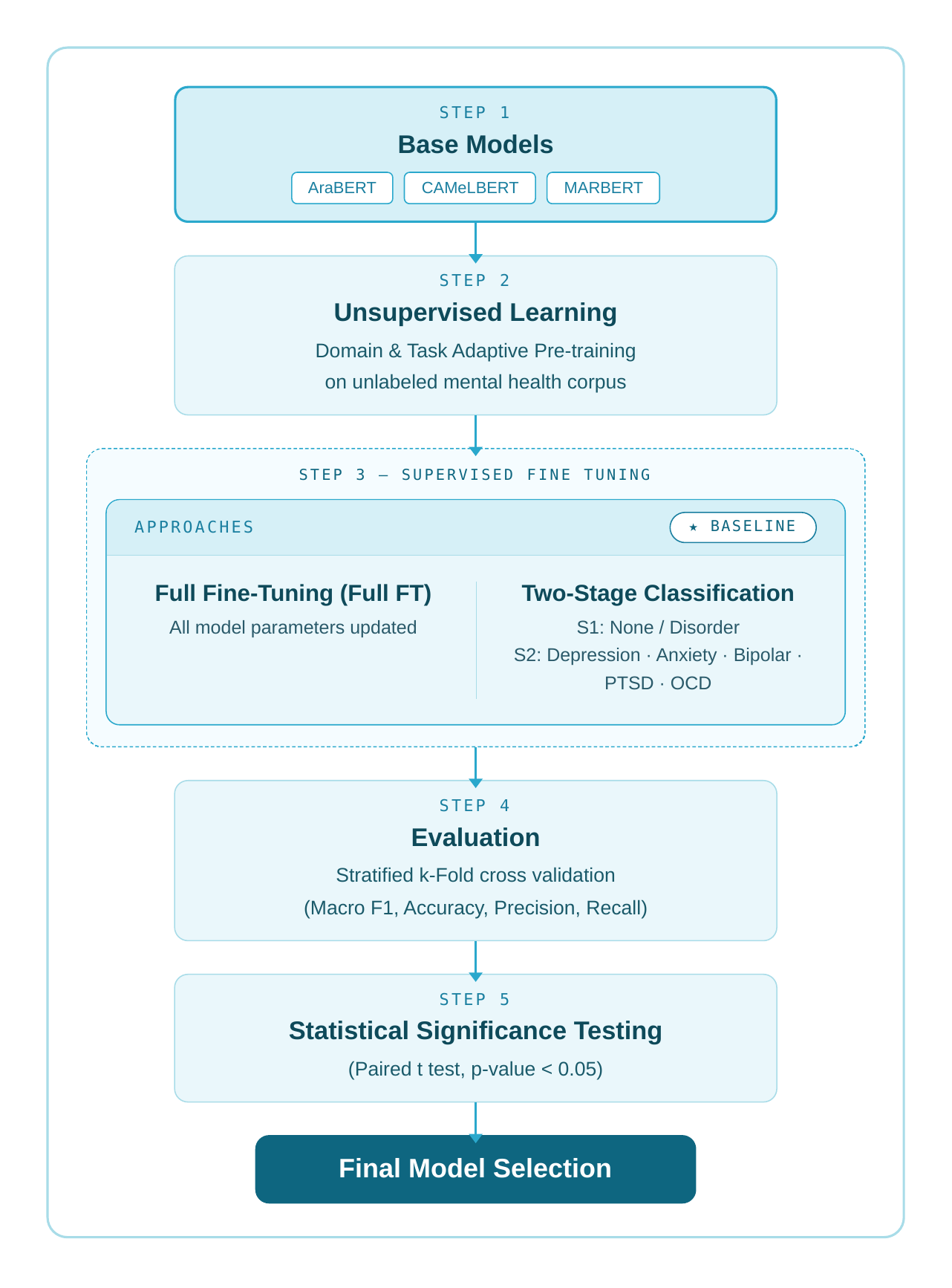}
    \caption{Overview of the proposed baseline framework for Arabic mental health disorder classification}
    \label{fig:phase1}
\end{figure}
\subsection {Phase 1: Baseline model selection}
The baseline selection framework follows five sequential steps to identify the best-performing 
Arabic pre-trained language model for mental health classification.

\subsubsection*{Step 1: Base Models}
Three Arabic pre-trained language models are used as base models: AraBERT, CAMeLBERT, and MARBERT. AraBERT is pre-trained on a large Arabic corpus, including Wikipedia, news articles, and web-crawled text, with Arabic-specific morphological segmentation~\cite{Abdul-mageed2021}. CAMeLBERT is pre-trained on diverse Arabic varieties including Modern Standard Arabic (MSA), dialectal, and classical Arabic, making it suitable for handling variation in writing styles~\cite{Inoue2021}. MARBERT is pre-trained on one billion tweets in both MSA and Arabic dialects, making it particularly suited for informal, colloquial language patterns found in social media mental health content~\cite{Abdul-mageed2021}. These models serve as the foundation for subsequent domain and task adaptive pre-training.

\subsubsection*{Step 2: Unsupervised Learning (Domain and Task 
Adaptive Pre-training)}
To better align the models with the linguistic characteristics of Arabic mental health discourse, 
two complementary pre-training strategies are applied on an unlabeled Arabic mental health corpus. DAPT exposes the model to domain-specific vocabulary and language patterns found in mental health texts, enabling it to develop a deeper understanding of clinical and psychological terminology~\cite{Ilias2024}. TAPT further specializes the model by pre-training on data that closely resembles the downstream classification task, narrowing the gap between general language understanding and task-specific requirements. Both strategies are implemented using Masked Language Modeling (MLM) with a dynamic masking rate of 15\%, where tokens are randomly masked during each training epoch to encourage the model to learn contextual representations~\cite{Gururangan2019}.
Formally, given an input sequence $\mathbf{x} = \{x_1, \dots, x_n\}$, a subset $M$ of token positions is randomly selected for masking. The model is trained to predict the original tokens at these positions by minimizing the MLM loss is defined as shown in Equation (4):
\begin{equation}
\mathcal{L}_{\text{MLM}} = - \sum_{i \in M} \log P(x_i \mid \mathbf{x}_{\setminus M})
\end{equation}
where $\mathbf{x}_{\setminus M}$ denotes the masked input sequence (i.e., the sequence where tokens in $M$ are replaced by the mask token), and $P(x_i \mid \mathbf{x}_{\setminus M})$ represents the probability assigned by the model to the correct token $x_i$ given its surrounding context. Model performance during pre-training is monitored using validation loss and perplexity as stopping criteria. Perplexity (PPL) is defined as shown in Equation (5):
\begin{equation}
\text{PPL} = \exp\left(\mathcal{L}_{\text{MLM}}\right)
\end{equation}

A lower perplexity indicates that the model assigns higher probability to the observed tokens, reflecting better adaptation to the target domain. Pre-training is halted at the point of optimal generalization to prevent overfitting to the unlabeled corpus.

\subsubsection {Step 3: Supervised Fine-Tuning (Baseline)}
Following pre-training, supervised fine-tuning is performed using Full Fine-Tuning (Full FT) combined with a Two-Stage Classification approach, applied consistently across all three base models. Full Fine-Tuning (Full FT) updates all model parameters during training, allowing the entire network to adapt to the classification task and maximizing the model's capacity to learn task-specific features. The two-stage classification approach is adopted to address class imbalance and inter-class confusion. These challenges arise when directly distinguishing between mental health disorders and the None class in a single step. In the first stage (S1), the model performs binary classification to determine whether the input text indicates the presence of a mental disorder or the absence of any disorder (None). In the second stage (S2), inputs classified as Disorder in S1 are passed to a second classifier that identifies the specific disorder type among Depression, Anxiety, Bipolar Disorder, PTSD, and OCD. This hierarchical decomposition reduces confusion between the None class and the disorder 
classes, improving overall classification performance.

\subsubsection {Step 4: Evaluation}
All three models are evaluated using stratified K-fold cross-validation to ensure that each fold maintains the original class distribution, providing reliable and unbiased performance estimates across imbalanced mental health categories. This evaluation strategy is applied to both the original pre-trained models and their domain-adapted counterparts, enabling a direct comparison that quantifies the contribution of DAPT and TAPT to classification performance. Performance is measured using macro F1-score, accuracy, macro precision, and macro recall, with macro F1 serving as the primary metric due to its sensitivity to performance across all classes regardless of class frequency. The measurement metrics used are:

Accuracy: measures the overall proportion of correctly classified instances, as demonstrated in Equation (6):

\begin{equation}
\text{Acc} = \frac{TP + TN}{TP + TN + FP + FN}
\end{equation}
Macro Precision: measures the average proportion of correctly 
predicted positive instances out of all predicted 
positives, averaged across all C classes, as shown in Equation (7):
\begin{equation}
P = \frac{1}{C} \sum_{c=1}^{C} \frac{TP_c}{TP_c + FP_c}
\end{equation}
Macro Recall: measures the average proportion of actual positive 
instances that were correctly identified, as shown in Equation (8):
\begin{equation}
R = \frac{1}{C} \sum_{c=1}^{C} \frac{TP_c}{TP_c + FN_c}
\end{equation}
Macro F1-score: is computed by averaging the 
per-class F1-scores across all $C$ classes, as shown in Equation (9):

\begin{equation}
\text{F1} = \frac{1}{C} \sum_{c=1}^{C} 
\frac{2 \cdot P_c \cdot R_c}{P_c + R_c}
\end{equation}

where $TP_c$, $FP_c$, and $FN_c$ denote the true positives, false positives, and false 
negatives for class $c$, respectively, and $C$ is the total number of classes.

\subsubsection*{Step 5: Statistical Significance Testing}
To ensure that observed performance differences between models are not attributable to random 
variation, a paired t-test is applied. The paired t-test compares the fold-level performance scores 
of two models across the same K-fold splits, testing the null hypothesis that no significant 
difference exists between them. Given two models with fold-level scores $\{a_1, \dots, a_K\}$ and 
$\{b_1, \dots, b_K\}$, the test statistic is computed as shown in equations (10) and (11):
\begin{equation}
t = \frac{\bar{d}}{s_d / \sqrt{K}}
\end{equation}
where:
\begin{equation}
\bar{d} = \frac{1}{K} \sum_{i=1}^{K} (a_i - b_i)
\end{equation}
and $s_d$ is the standard deviation of the 
pairwise differences $d_i = a_i - b_i$, and 
$K$ is the number of folds. A p-value threshold 
of 0.05 is used, where $p < 0.05$ indicates a 
statistically significant difference. The 
best-performing model identified through this 
process is selected as the base model for the 
proposed classification framework in the 
subsequent phase.

\subsection{Phase 2: Proposed classification framework}

\begin{figure}[!htbp]
  \centering
   \includegraphics[width=0.5\linewidth]{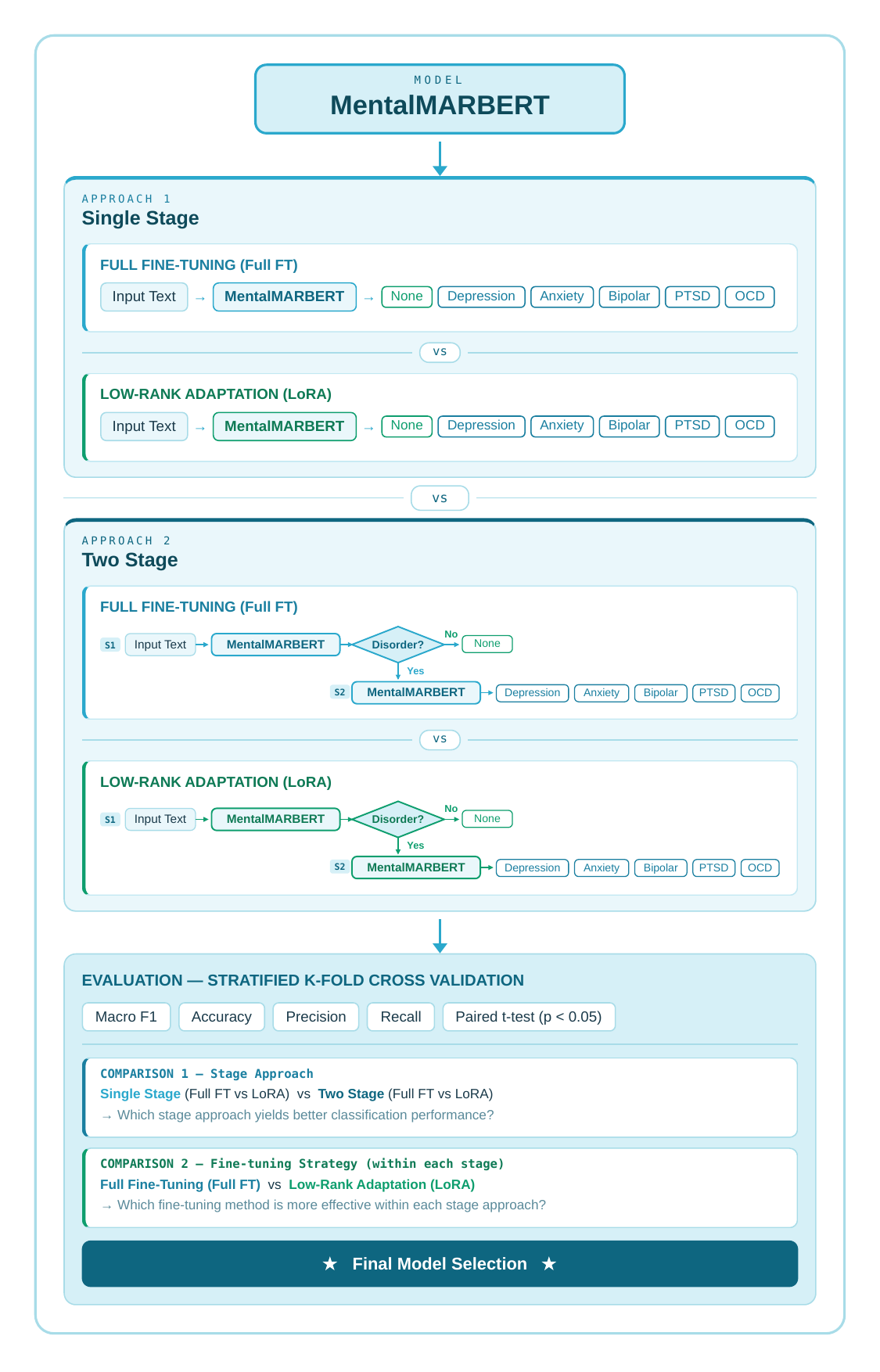}
   \caption{Experimental design of the proposed MentalMARBERT model, illustrating the comparison between single-stage and two-stage classification using full FT and LoRA.}
  \label{fig:phase2}
\end{figure}

Figure 5 presents the proposed classification framework built upon MentalMARBERT, the best-performing backbone model identified in Phase 1. This phase systematically evaluates four experimental configurations 
derived from the combination of two classification architectures and two fine-tuning strategies.
The first architecture, single-stage classification, performs direct six-class prediction. The input text is processed by MentalMARBERT and classified into one of the predefined categories: None, Depression, Anxiety, Bipolar Disorder, PTSD, or OCD in a single forward pass. This approach treats the task as a standard multi-class classification problem.
The second architecture, two-stage classification, adopts a hierarchical design to mitigate confusion between the None class and disorder-related classes. In Stage 1 (S1), the model performs binary classification to determine whether the input text reflects the presence of a mental disorder (Disorder) or its absence (None). In Stage 2 (S2), only texts predicted as Disorder in S1 are forwarded to a second classifier that identifies the specific disorder type among Depression, Anxiety, Bipolar Disorder, PTSD, and OCD. Each stage is trained independently using the cross-entropy loss function. For both architectural approaches, two fine-tuning strategies are investigated. In Full Fine-Tuning (Full FT), all model parameters are updated during supervised training, enabling complete adaptation of the backbone model to the downstream task. In contrast, Low-Rank Adaptation (LoRA) introduces trainable low-rank matrices into the attention layers while freezing the original backbone parameters~\cite{Hu2022}. This parameter-efficient approach substantially reduces the number of trainable parameters while maintaining competitive classification performance. All four configurations are evaluated using stratified K-fold cross-validation to preserve class distribution across folds. Performance is measured using accuracy, precision, recall, and macro F1-score, with macro F1 serving as the primary evaluation metric given the inherent class imbalance in the dataset. Two comparative analyses are conducted. First, single-stage and two-stage architectures are compared to determine which classification design better handles multi-class mental health detection. Second, within each architectural setup, Full FT and LoRA are compared to assess the effectiveness of parameter-efficient adaptation. Statistical significance is assessed using a paired t-test with a threshold of p < 0.05. The configuration achieving the highest statistically significant performance is selected as the final model.

\section{Experimental setup}\label{sec5}
This section presents a comprehensive description of the experimental configurations utilized to evaluate the framework, 
which are explained in section IV, Phase 1, and Phase 2. The experimental design comprises four principal components: the computational environment, hyperparameter settings for domain-adaptive pre-training, fine-tuning configurations for both baseline and proposed setups, and the evaluation protocol applied across all experiments. All experiments were executed on Google Colab Pro using an NVIDIA A100 GPU. The implementation was based on PyTorch 2.10.0 and the Hugging Face Transformers library 5.0.0

\subsection{Phase 1: Baseline Configuration}

Domain and Task Adaptive Pre-training (DAPT and TAPT) was performed using Masked Language Modeling (MLM) with a dynamic masking rate of 15\%, following the standard BERT pretraining setup ~\cite{devlin2019bert} with validation loss and perplexity monitored throughout training. Hyperparameter optimization was conducted in two stages. First, each model was evaluated using its default hyperparameters as a baseline configuration. Subsequently, Optuna was employed to perform automated hyperparameter search, tuning each model independently across the following search ranges: 
learning rate (1e-5 to 5e-5), batch size (8 to 128), and number of epochs (3 to 15). The configuration 
yielding the lowest validation loss was selected for each model, as summarized in Table 3.

\begin{table}[H]
\centering
\caption{Domain and Task Adaptive Pre-Training Hyperparameters}
\label{tab:dapt_hparams}
\begin{tabular}{@{}lllll@{}} 
\toprule
\textbf{Model} & \textbf{LR} & \textbf{Batch} & \textbf{Epochs} & \textbf{Optimizer} \\
\midrule
AraBERTv0.2 - twitter & $2\times10^{-5}$     & 16 & 10 & AdamW \\
MARBERTv2             & $2.60\times10^{-5}$  & 8  & 5  & Adafactor \\
CAMeLBERT             & $4.93\times10^{-5}$  & 32 & 5  & AdamW \\
\bottomrule
\end{tabular}
\end{table}
Following pre-training, supervised fine-tuning was performed using the two-stage classification approach 
with full fine-tuning applied consistently across all base models with DAPT. The fine-tuning hyperparameters were configured independently for each model, and the configuration yielding the lowest validation loss was selected, as detailed in Table 4.


\begin{table}[H]
\centering
\caption{Fine-Tuning Configuration (Two-Stage with Full FT)}
\label{tab:ft_comparison}
\begin{tabular}{@{}llllll@{}} 
\toprule
\textbf{Model} & \textbf{Setting} & \textbf{Strategy} & \textbf{LR} & \textbf{Batch} & \textbf{Epochs} \\
\midrule
AraBERTv0.2-twitter & Base & Full FT & $2\times10^{-5}$ & 32 & 3 \\
MARBERTv2           & Base & Full FT & $1\times10^{-5}$ & 32 & 3 \\
CAMeLBERT           & Base & Full FT & $2\times10^{-5}$ & 32 & 3 \\
\midrule
MentalAraBERT       & DAPT & Full FT & $2\times10^{-5}$ & 32 & 3 \\
MentalMARBERT       & DAPT & Full FT & $1\times10^{-5}$ & 32 & 3 \\
MentalCAMeLBERT     & DAPT & Full FT & $2\times10^{-5}$ & 32 & 3 \\
\bottomrule
\end{tabular}
\end{table}

\subsection{Phase 2: Proposed classification framework}
MentalMARBERT was selected as the backbone model for Phase 2 following the evaluation conducted in Phase 1.
In Phase 2, MentalMARBERT was evaluated across four experimental configurations derived from combining two classification approaches with two fine-tuning strategies as shown in Table \ref{tab:phase2_finetune_hparams}. For full fine-tuning configurations, all model parameters were updated during training. For LoRA-based configurations, trainable low-rank matrices were introduced into the attention layers of MentalMARBERT while keeping the base model parameters frozen.
\begin{table}[h]
\centering
\caption{Phase 2: Fine-Tuning Hyperparameters}
\label{tab:phase2_finetune_hparams}
\begin{tabular}{@{}llllll@{}}
\toprule
\textbf{Configuration} & \textbf{Approach} & \textbf{Strategy} & \textbf{LR} & \textbf{Batch} & \textbf{Epochs} \\
\midrule
Config 1 & Single-Stage & Full FT & $1\times10^{-5}$ & 32 & 3 \\
Config 2 & Single-Stage & LoRA    & $1\times10^{-5}$ & 32 & 3 \\
Config 3 & Two-Stage    & Full FT & $1\times10^{-5}$ & 32 & 3 \\
Config 4 & Two-Stage    & LoRA    & $1\times10^{-5}$ & 32 & 3 \\
\bottomrule
\end{tabular}
\footnotetext{Note: All Phase 2 fine-tuning experiments were evaluated using stratified 5-fold cross-validation ($k=5$).}
\end{table}

Hyperparameter optimization for the fine-tuning configurations was performed using Optuna ~\cite{Akiba2019}, an open-source hyperparameter optimization framework that automates the search for optimal model configurations, with 50 trials. The search included learning rate, batch size, number of training epochs, and LoRA-specific parameters (rank, alpha, and dropout). Table~\ref{tab:optuna_search} presents the hyperparameter search space explored during Optuna optimization, where the learning rate was searched within the range of $1\times10^{-5}$ to $5\times10^{-5}$, the batch size ranged from 8 to 128, and the number of training epochs was varied between 3 and 10.

\begin{table}[h]
\centering
\caption{Optuna Hyperparameter Search Space}
\label{tab:optuna_search}
\renewcommand{\arraystretch}{1.2}
\setlength{\tabcolsep}{6pt}
\begin{tabular}{|l|c|}
\hline
\textbf{Hyperparameter} & \textbf{Range} \\
\hline
Learning rate & $1\times10^{-5}$ to $5\times10^{-5}$ \\
\hline
Batch size & 8 to 128 \\
\hline
Number of epochs & 3 to 10 \\
\hline
\end{tabular}
\end{table}

Table~\ref{tab:lora_explored} presents the explored LoRA configurations, where the rank (r) was varied across three values (8, 16, and 32), the alpha parameter was set correspondingly to 16, 32, and 64, and the dropout rate ranged from 0.01 to 0.10.

\begin{table}[h]
\centering
\caption{Explored LoRA Configurations}
\label{tab:lora_explored}
\renewcommand{\arraystretch}{1.2}
\begin{tabular}{|c|c|c|}
\hline
\textbf{Rank ($r$)} & \textbf{Alpha} & \textbf{Dropout} \\
\hline
8  & 16 & 0.01 \\
\hline
16 & 32 & 0.05 \\
\hline
32 & 64 & 0.10 \\
\hline
\end{tabular}
\end{table}
To ensure a fair comparison, the same evaluation protocol used in Phase 1 was applied in Phase 2. All configurations were evaluated using stratified 5-fold cross-validation (k = 5). Performance was assessed using accuracy, precision, recall, and macro-F1, with statistical significance verified using a paired two-tailed t-test at a significance level of $\alpha = 0.05$.

\section{Results and Discussion}\label{sec12}
This section presents the experimental results obtained across both phases of the proposed framework, followed by a discussion of the key findings. Phase 1 reports the impact of domain and task adaptive pre-training on the three base models and identifies the best-performing backbone model. Phase 2 presents the comparative evaluation of the four experimental configurations combining single-stage and two-stage classification with full fine-tuning and LoRA.

\subsection{Baseline Model Selection Results}
Table 8 presents the validation loss and perplexity obtained after domain-adaptive pre-training (DAPT) on the collected Arabic mental health corpus. Among the evaluated models, AraBERT achieved the lowest validation loss (2.19) and perplexity (8.96), indicating better adaptation to the target domain compared to MARBERTv2 and CAMeLBERT.

MARBERTv2 recorded the highest validation loss (3.04) and perplexity (21.07), suggesting weaker alignment with the mental health domain under the same pre-training configuration. On the other hand, CAMeLBERT showed intermediate performance, with a validation loss of 2.76 and perplexity of 15.87.

Overall, the results indicate that AraBERT benefited most from continued pre-training on domain-specific data, demonstrating stronger contextual adaptation to mental health–related linguistic patterns. Lower perplexity values reflect improved predictive confidence of the masked language modeling objective on validation data.

\begin{table}[h]
\centering
\caption{Continued Pre-Training (MLM) Results}
\label{tab:pretraining_results}
\begin{tabular}{@{}lcc@{}}
\toprule
\textbf{Model} & \textbf{Validation Loss} & \textbf{Perplexity} \\
\midrule
AraBERTv0.2-twitter & 2.19 & 8.96  \\
MARBERTv2           & 3.04 & 21.07 \\
CAMeLBERT           & 2.76 & 15.87 \\
\bottomrule
\end{tabular}
\end{table}

Table~\ref{tab:phase1_classification} presents the comparative results of the three base models before and after domain and task adaptive pretraining, evaluated under the two-stage full fine-tuning setup using stratified 5-fold cross-validation. In addition, each model was compared to its domain-adaptive pre-training version as follows:

\subsubsection{AraBERT vs MentalAraBERT}
The results indicate that domain and task adaptive pre-training does not yield statistically significant improvements for AraBERT across any evaluation metric. Although a marginal improvement was observed in accuracy ($+0.0056$) and recall ($+0.0037$), none of these differences reached statistical significance, suggesting that AraBERT's original pre-training on formal Arabic text does not sufficiently align with the informal and colloquial nature of Arabic mental health discourse.

\subsubsection{CAMeLBERT vs MentalCAMeLBERT}
For CAMeLBERT, domain and task adaptive pre-training yielded a statistically significant improvement only in Recall ($p = 0.0028$, $+0.0152$), indicating that MentalCAMeLBERT became more sensitive to detecting disorder-related instances after pre-training. However, no statistically significant improvement was observed in Macro-F1 ($p = 0.453$), and precision slightly declined ($-0.0013$), suggesting a precision--recall trade-off rather than a consistent overall performance gain.

\subsubsection{MARBERT vs MentalMARBERT}
MentalMARBERT achieved the most consistent and robust improvements following domain and task adaptive pre-training. Statistically significant gains were observed in both accuracy ($p < 0.001$, $+0.0103$) and macro-F1 ($p = 0.0309$, $+0.0080$), confirming that pre-training on Arabic mental health data meaningfully enhanced MARBERT's ability to classify mental health disorders. This is attributed to MARBERT's original pre-training on Twitter data, which closely aligns with the informal and colloquial language patterns found in the mental health corpus used for adaptation.

\begin{table}[h]
\centering
\caption{Phase 1: Comparative Classification Results Before and After Domain Adaptation}
\label{tab:phase1_classification}
\begin{tabular}{@{}lllllll@{}}
\toprule
\textbf{Model} & \textbf{Metric} & \textbf{Before} & \textbf{After} & \textbf{Mean Diff} & \textbf{$p$-value} & \textbf{Sig} \\
\midrule
AraBERT & Accuracy  & 0.8666 & 0.8722 & 0.0056  & 0.159 & $\times$ \\
MentalAraBERT & Macro Precision & 0.8456 & 0.8471 & 0.0015  & 0.580 & $\times$ \\
& Macro Recall    & 0.8613 & 0.8651 & 0.0037  & 0.230 & $\times$ \\
& Macro-F1  & 0.8524 & 0.8520 & $-0.0004$ & 0.683 & $\times$ \\
\midrule
CAMeLBERT & Accuracy  & 0.8668 & 0.8669 & 0.0001  & 0.788 & $\times$ \\
MentalCAMeL & Macro Precision & 0.8459 & 0.8446 & $-0.0013$ & 0.349 & $\times$ \\
&Macro  Recall    & 0.8622 & 0.8774 & 0.0152  & \textbf{0.0028} & $\checkmark$ \\
& Macro-F1  & 0.8530 & 0.8525 & $-0.0005$ & 0.453 & $\times$ \\
\midrule
MARBERT & Accuracy  & 0.8675 & 0.8778 & 0.0103  & \textbf{0.001} & $\checkmark$ \\
MentalMARBERT & Macro Precision & 0.8467 & 0.8473 & 0.0006  & 0.529 & $\times$ \\
& Macro Recall    & 0.8628 & 0.8651 & 0.0024  & 0.071 & $\times$ \\
& Macro-F1  & 0.8537 & 0.8617 & 0.0080  & \textbf{0.0309} & $\checkmark$ \\
\bottomrule
\end{tabular}
\footnotetext{Note: $\checkmark$ indicates statistically significant improvement at $\alpha = 0.05$, while $\times$ indicates non-significant results.}
\end{table}
\subsubsection {Model selection}
Model selection was based on the mean Macro-F1 score as the primary criterion, with fold-level stability across the 5-fold cross-validation splits used as a secondary criterion to ensure that the selected model generalizes consistently rather than performing well on specific folds only. Among all evaluated models, MentalMARBERT achieved the highest Macro-F1 score of 0.8617, representing an improvement of +0.0080 over its base model, MARBERT (0.8537). Critically, this improvement was statistically significant (p = 0.0309), confirming that the observed gain is not due to random variation. In addition, MentalMARBERT achieved a statistically significant improvement in accuracy (p < 0.001, +0.0103), further reinforcing the robustness of its performance. In contrast, MentalAraBERT and MentalCAMeLBERT failed to demonstrate consistent statistically significant improvements across the primary evaluation metrics. MentalAraBERT showed no significant improvement in any metric, while MentalCAMeLBERT achieved significance only in recall, without a corresponding improvement in Macro-F1. Therefore, MentalMARBERT was selected as the backbone model for Phase 2, as it demonstrated the most consistent, statistically significant, and generalizable performance gains following domain and task adaptive pre-training.
\subsection {Phase 2: Proposed classification framework Results}
The following results are obtained from the four configurations evaluated on MentalMARBERT across 
two comparative analyses. The first analysis compares fine-tuning strategies (full FT vs LoRA) within each architecture type, while the second analysis compares architectural designs (Single Stage vs Two Stage) within each fine-tuning strategy. All results are 
reported as mean ± standard deviation across 5-fold cross-validation splits.

\subsubsection {Single-Stage: Full FT vs LoRA} 
Under the single-stage architecture, full fine-tuning achieved statistically significant improvements over LoRA in Accuracy (p = 0.0207, +0.0033), precision (p = 0.0029, +0.0051), and macro-F1 (p = 0.0189, +0.0043). No statistically significant difference was observed in recall (p = 0.1807), suggesting that both strategies perform comparably in identifying disorder-related instances under the single-stage, as detailed in Table 10.
\begin{table}[h]
\centering
\caption{Single-Stage: Full FT vs LoRA Performance Comparison}
\label{tab:Single_stage}
\begin{tabular}{@{}llllll@{}}
\toprule
\textbf{Metric} & \textbf{Full FT (Mean $\pm$ SD)} & \textbf{LoRA (Mean $\pm$ SD)} & \textbf{Mean Diff} & \textbf{$p$-value} & \textbf{Sig} \\
\midrule
Accuracy  & \textbf{0.8703 $\pm$ 0.0044} & 0.8670 $\pm$ 0.0033 & +0.0033 & 0.0207 & $\checkmark$ \\
Macro Precision & \textbf{0.8490 $\pm$ 0.0040} & 0.8439 $\pm$ 0.0036 & +0.0051 & 0.0029 & $\checkmark$ \\
Macro Recall    & 0.8658 $\pm$ 0.0089 & 0.8633 $\pm$ 0.0075 & +0.0026 & 0.1807 & $\times$ \\
Macro-F1  & \textbf{0.8564 $\pm$ 0.0058} & 0.8521 $\pm$ 0.0049 & +0.0043 & 0.0189 & $\checkmark$ \\
\bottomrule
\end{tabular}
\end{table}
\subsubsection {Two-Stage: Full FT vs LoRA}
Within the two-stage architecture, full fine-tuning consistently outperformed LoRA 
across all primary metrics. Statistically significant improvements were observed in accuracy (p < 0.01, +0.0124), precision 
(p < 0.01, +0.0052), and macro-F1 (p < 0.01, +0.0111), indicating that full parameter updates are particularly beneficial 
for the hierarchical classification structure. As in the Single-Stage setting, no significant 
difference was observed in Recall (p = 0.11), as shown in Table 11.

\begin{table}[h]
\centering
\caption{Two-Stage: Full FT vs LoRA Performance Comparison}
\label{tab:Two_stage}
\begin{tabular}{@{}llllll@{}}
\toprule
\textbf{Metric} & \textbf{Full FT (Mean $\pm$ SD)} & \textbf{LoRA (Mean $\pm$ SD)} & \textbf{Mean Diff} & \textbf{$p$-value} & \textbf{Sig} \\
\midrule
Accuracy  & \textbf{0.8778 $\pm$ 0.0011} & 0.8654 $\pm$ 0.0023 & +0.0124 & 0.01 & $\checkmark$ \\
Macro Precision & \textbf{0.8473 $\pm$ 0.0033} & 0.8421 $\pm$ 0.0019 & +0.0052 & 0.01 & $\checkmark$ \\
Macro Recall    & 0.8651 $\pm$ 0.0039 & 0.8625 $\pm$ 0.0042 & +0.0026 & 0.11 & $\times$ \\
Macro-F1  & \textbf{0.8617 $\pm$ 0.0062} & 0.8506 $\pm$ 0.0013 & +0.0111 & 0.01 & $\checkmark$ \\
\bottomrule
\end{tabular}
\end{table}

\subsubsection {Architecture Comparison (Full FT)
Single-Stage vs Two-Stage} When full fine-tuning is applied, the two-stage 
architecture achieves statistically significant improvements over Single-Stage in both accuracy 
(p < 0.05, +0.0075) and macro-F1 (p < 0.05, +0.0053), as summarized in Table 12. No significant differences were observed 
in precision or recall, suggesting that the hierarchical decomposition primarily benefits 
overall classification accuracy and balanced class performance rather than class-specific precision or recall.
\begin{table}[h]
\centering
\caption{Architecture Comparison (Full FT): Single-Stage vs Two-Stage Performance}
\label{tab:architecture_comparison}
\begin{tabular}{@{}llllll@{}}
\toprule
\textbf{Metric} & \textbf{Single-Stage} & \textbf{Two-Stage} & \textbf{Mean Diff} & \textbf{$p$-value} & \textbf{Sig} \\
\midrule
Accuracy  & 0.8703 $\pm$ 0.0044 & \textbf{0.8778 $\pm$ 0.0011} & $-0.0075$ & 0.05 & $\checkmark$ \\
Macro Precision & \textbf{0.8490 $\pm$ 0.0040} & 0.8473 $\pm$ 0.0033 & +0.0017 & 0.21 & $\times$ \\
Macro Recall    & \textbf{0.8658 $\pm$ 0.0089} & 0.8651 $\pm$ 0.0039 & +0.0007 & 0.74 & $\times$ \\
Macro-F1  & 0.8564 $\pm$ 0.0058 & \textbf{0.8617 $\pm$ 0.0062} & $-0.0053$ & 0.05 & $\checkmark$ \\
\bottomrule
\end{tabular}
\end{table}
\subsubsection {Architecture Comparison (LoRA):
Single-Stage vs Two-Stage}
Under LoRA-based fine-tuning, no statistically significant differences were observed between single-stage and two-stage architectures across any metric, with p-values ranging from 0.08 to 0.31 (see Table 13). This suggests that when the model's parameters are largely frozen, the architectural design has a limited impact on classification performance.

\begin{table}[H]
\centering
\caption{Architecture Comparison (LoRA): Single-Stage vs Two-Stage Performance}
\label{tab:architecture_comparison_lora}
\begin{tabular}{@{}llllll@{}}
\toprule
\textbf{Metric} & \textbf{Single-Stage} & \textbf{Two-Stage} & \textbf{Mean Diff} & \textbf{$p$-value} & \textbf{Sig} \\
\midrule
Accuracy  & \textbf{0.8670 $\pm$ 0.0033} & 0.8654 $\pm$ 0.0023 & +0.0016 & 0.09 & $\times$ \\
Macro Precision & \textbf{0.8439 $\pm$ 0.0036} & 0.8421 $\pm$ 0.0019 & +0.0018 & 0.12 & $\times$ \\
Macro Recall    & \textbf{0.8633 $\pm$ 0.0075} & 0.8625 $\pm$ 0.0042 & +0.0008 & 0.31 & $\times$ \\
Macro-F1  & \textbf{0.8521 $\pm$ 0.0049} & 0.8506 $\pm$ 0.0013 & +0.0015 & 0.08 & $\times$ \\
\bottomrule
\end{tabular}
\end{table}
\FloatBarrier

\subsubsection {Final Model Selection}
Based on the comparative evaluation, the two-stage architecture combined with full fine-tuning 
achieved the highest overall performance, with a macro-F1 of 0.8617 and accuracy of 0.8778. 
This configuration demonstrated statistically significant superiority over all other evaluated 
configurations and was therefore selected as the final model.

\section{Conclusion}\label{sec13}

This paper presented a comprehensive framework for Arabic mental health text classification, structured across two sequential phases. Phase 1 focused on identifying the most suitable backbone model through domain and task adaptive pre-training of three Arabic pre-trained language models, AraBERT, CAMeLBERT, and MARBERT on an unlabeled Arabic mental health corpus. Phase 2 evaluated four experimental configurations on the selected backbone model, MentalMARBERT, by combining Single-Stage and Two-Stage classification architectures with Full Fine-Tuning and Low-Rank Adaptation (LoRA).
The experimental results demonstrated that domain and task adaptive pre-training does not uniformly benefit all models. While AraBERT showed no statistically significant improvement and CAMeLBERT benefited only in recall, MentalMARBERT achieved consistent and statistically significant gains in both accuracy and macro-F1, confirming its suitability as the backbone model for the proposed framework. This finding highlights the importance of alignment between a model's original pre-training data and the target domain, as MARBERT's Twitter-based pre-training closely mirrors the informal language patterns found in Arabic mental health discourse.
In Phase 2, the two-stage architecture combined with full fine-tuning emerged as the best-performing configuration, achieving a macro-F1 of 0.8617 and accuracy of 0.8778 with statistically significant improvements over all other configurations. The hierarchical decomposition of the classification task effectively reduced inter-class confusion between the None class and disorder-related categories, validating the motivation behind the two-stage design. Furthermore, full fine-tuning consistently outperformed LoRA across both architectures, although LoRA demonstrated competitive performance under the single-stage setup, suggesting its potential as a parameter-efficient alternative in resource-constrained settings. These findings contribute to the growing body of research on Arabic NLP for mental health applications, demonstrating that task-specific pre-training combined with hierarchical classification and full parameter adaptation yields robust and reliable performance for Arabic mental health disorder detection.

\bibliographystyle{unsrt}
\bibliography{references}

\end{document}